# Khayyam Offline Persian Handwriting Dataset

*Pourya Jafarzadeh*
p.jafarzadeh@tosantechno.com

*Padideh Choobdar*
p.choobdar@gmail.com

Vahid Mohammadi Safarzadeh
vahid.msafarzadeh@scu.ac.ir

**Abstract**

Handwriting analysis is still an important application in machine learning. A basic requirement for any handwriting recognition application is the availability of comprehensive datasets. Standard labelled datasets play a significant role in training and evaluating learning algorithms. In this paper, we present the Khayyam dataset as another large unconstrained handwriting dataset for elements (words, sentences, letters, digits) of the Persian language. We intentionally concentrated on collecting Persian word samples which are rare in the currently available datasets. Khayyam's dataset contains 44000 words, 60000 letters, and 6000 digits. Moreover, the forms were filled out by 400 native Persian writers. To show the applicability of the dataset, machine learning algorithms are trained on the digits, letters, and word data and results are reported. This dataset is available for research and academic use.

## 1. Introduction

Offline handwriting recognition is the process of converting the scanned (photographed) image of a handwritten script into an editable format. Since in machine learning we need a reasonable amount of data to learn discriminative features, it is always worth collecting more data from various sources and from a variety of distributions [1].

The task of Persian handwriting recognition has several specific complexities including variable forms of each letter in different positions in a word, various shapes for a single letter, the cursive nature of handwriting styles, the existence of letters having ascenders and dots. Valuable efforts have been made previously for gathering Persian and Arabic handwritten digits and letters. However, the amount of data for words and sentences especially Persian words was not enough. In addition, there is not a comprehensive Persian dataset with a perfect ground truth label for researchers.

Constructing a handwriting dataset is a time-consuming procedure due the challenges of collecting data from many writers as well as the labelling procedures [2]. For a handwriting dataset to be of scientific value, it should cover many varieties of digits, numbers, letters, words, and sentences.

A lot of efforts have been made to provide handwriting databases in many languages [3, 4, 5, 6, 7]. However, the amount of data on the Persian language is still low. For this reason, we introduce the Khayyam database, a comprehensive handwriting collection of letters, numbers, words, sentences and arithmetic signs written by Persian writers. In the following paragraphs, we introduced the few datasets for the Persian language that are already available.

One of the most recent examples is reported in [8] in which the authors gathered handwritten samples from 250 males and 250 females. All types of data are provided including digits, numbers, alphabet letters, words (city names, people's names), texts, and arithmetic signs. Unfortunately, the dataset lacks an organised structure. For example, the number of names of cities for each city is different from others and this is so for people's names. More importantly, the dataset did not have an easy-access labelling structure for the data, particularly for the numbers and words which we required. The data are just placed in directories with no pattern. For the sake of completeness of the comparisons in this work, we generated some scripts to organise parts of the data.

In [9], a dataset is provided for specific usages in writer identification. For this purpose, 200 forms were collected from native writers of different ages and genders. Since the main goal of this work was to recognize the writing style, the forms were designed accordingly and contained places for writing paragraphs of texts in Persian and English. The dataset only contains sentences.

The researchers in [10] provided images of filled forms and bank checks with signatures alongside their ground truth. Several forms including the ones for signatures, Persian numbers (used in the Iranian banking system), and dates were collected. Since the data is for special use only, it cannot be considered a comprehensive dataset for general purposes.

The authors in [11] collected documents from a national newspaper (1996 to 2002) that were written by many authors from a variety of backgrounds and covered a range of different topics, each with a credible size of data. The data of the dataset are in the printed form, the researchers of [12] used the dataset to develop a printed Persian OCR system.

Inspired by the IFN/ENIT database for Arabic handwritten texts [7], the authors of [13] collected 7271 binary images of 1080 Iranian handwritten city names in Persian. In addition to being proper only for a special usage, it seems that the dataset is not supported by the authors anymore so we could not access it.

In [14] a large dataset of Persian digits is extracted from 12000 registration forms and labelled. In [15], a dataset of Persian characters, digits, and numbers is provided using forms that were written by 175 writers. Despite the considerable amount of data, these datasets lack labelled handwritten words.

The PHTD dataset presented in [16] contains only handwritten sentences extracted from 140 documents in three contexts: history, school dictations, and general texts. Forty individuals filled the documents.

In [6] a dataset of Persian, Bangla, Oriya, and Kannada (PBOK) is introduced which contains 707 text pages that were filled by 436 writers. The dataset includes 12565 text lines, 104541 words, and 423980 characters. It has two types of ground truths; one is based on the pixel information and the other is based on the image's content. The Persian data in this paper are the same as those in [16].

The Haft dataset, reported in [17], contains 1800 grayscale images collected from 600 writers. Each form has eight lines, and every writer filled each form three times. The writers, themselves, selected sentences without any restrictions. The handwritten words in this work were also not isolated and labelled.

The Iranshahr dataset [18] includes 19,583 images of 503 Iranian city names. There are at least 20 samples of each class with a variety of handwriting styles in the dataset. The dataset is useful for specific applications like reading post-address labels.

The dataset introduced in [19], IAUT/PHVN, was gathered from 380 writers. It has four subsets A, B, C and D with 34200 images with ground truth in aggregate. Despite its neatly organised structure, its domain is only limited to city names identification.

The PHCWT dataset introduced in [20] consists of 51200, 3600, and 400 images of characters, words, and texts, respectively. The data were collected from 400 people and each person filled in three forms. The number of words is relatively small, and it lacks the numerical data.

Despite the listed efforts, there is still a need to have more comprehensive labelled handwriting datasets for the Persian language, due to the complexity of the style and variability in the words and character forms.

Considering the shortcomings of the available datasets, this research introduces another Persian handwriting dataset.

This paper contains three parts. First, we discuss a few main differences and similarities between the Persian and Arabic languages having almost similar written forms and the challenges of text recognition tasks in both languages. Next, a detailed introduction to our database is explained. Eventually, to represent usability of the dataset, classification results of learning methods, K-Nearest Neighbours (KNN) and Support Vector Machines (SVM), and Convolutional Neural Networks-Recurrent Neural Networks (CNN-RNN), adjusted and trained by the database are reported.

## 2. Persian and Arabic Handwriting Differences

Both the Persian and Arabic languages use similar alphabets. However, even in writing, the forms are different. Therefore, datasets generated for one language, cannot be used for another. As you will see in the following list, the scripts written in both languages have some major differences despite their similarity.

- The Persian language has four more letters than Arabic: "پ" (P-e), "ژ" (Zh-e), "گ" (G-e), and "چ" (Che).
- In a common Persian script, it is unlikely to see words with "ة" (Ta Tanis which sounds similar to T). Instead, it converts to "ه" (He) in loan words from Arabic.
- The usage of the signs Tanvin, Saken (to represent a stop), Tashdid (to intensify the spelling of a letter) and Hamzeh (which sounds like Aa, see Fig. 1) is exceedingly rare in the Persian script. These signs are specific to Arabic and when they will be dropped out in loan words in Persian texts.

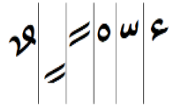

Fig. 1. From left to right: Oon, En, An (Three forms of Tanvin), Saken, Tashdid and Hamzeh.

- Vowels are not written in Persian texts except in places where the writer tries to eliminate any potential ambiguity, or in the first year of school (Fig. 2).

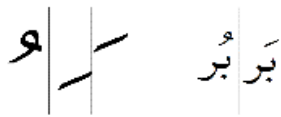

Fig. 2. a) The Persian vowels: from left to right sounds Oo, Ee, Aa, respectively. b) Using the vowels to eliminate ambiguity: left: Bor (meaning cut), right: Bar (meaning take or chest).

- The patterns of making plurals are not the same the two languages. In Persian, the preference is using "ها" (Ha), "ان" (Aan), or "جات" (Jat) at the end of the words to become plural. However, in Arabic writers use "ات" (Aat), "ین" (Ein), or "ون" (Oun) for similar purposes.
- Loan words from Arabic are not usually written in their original forms in Persian. In Arabic, the combination of the two letters "ا" (Aa) and "ل" (La) which makes "ال" (Al) influences the form of the letters following them. Using "Al" correctly at the start of nouns is crucial in Arabic. However, in Persian, it is generally wrong to use it for loan words except in some rare compounds. Therefore, every learning model trained on Arabic words datasets will learn a lot of words beginning with "Al" which does not apply to Persian applications.

Despite the differences between the two languages' handwriting styles, there are the same challenges in applications aiming to analyze texts written in Persian or Arabic.

## 2.1. Challenges of Persian/Arabic Handwriting Recognition

The process of Persian/Arabic handwriting recognition requires dealing with several complexities. Here we take the most challenging ones into account.

- **Letters with Similar Pattern:** Among Persian letters, there are letters with very similar main shapes as categorized in Table 1. This causes misclassification by character recognition methods.

- **Misplaced Dots:** Another issue is the position of the secondary components (particularly dots) in a word. Sometimes the writers put dots or ascenders not in their correct positions, because of fast writing, which mislead the pattern recognition methods (Fig. 3).

Table 1. Different forms of the Persian Letters categorized based on their similarity in shape.

| Medial Form | Initial Form |
|---|---|
| ب ـ پ ـ ت ـ ث ـ ن ـ يـ ـ ـفـ ـ ـقـ ـ ـعـ ـ ـغـ | بـ ـ پـ ـ تـ ـ ثـ ـ نـ ـ يـ ـ فـ ـ قـ |
| ـحـ ـ ـجـ ـ ـچـ ـ ـخـ | حـ ـ جـ ـ چـ ـ خـ |
| ـسـ ـ ـشـ ـ ـصـ ـ ـضـ ـ ـطـ ـ ـظـ | سـ ـ شـ ـ صـ ـ ضـ ـ طـ ـ ظـ |
|  | عـ ـ غـ ـ فـ ـ قـ |

| Final Form | Isolated Form |
|---|---|
| ـب ـ ـپ ـ ـت ـ ـث ـ ـف | ب ـ پ ـ ت ـ ث ـ ف |
| ـح ـ ـج ـ ـچ ـ ـخ ـ ـع ـ ـغ | ح ـ ج ـ چ ـ خ ـ ع ـ غ |
| ـد ـ ـذ ـ ـر ـ ـز ـ ـژ | د ـ ذ ـ ر ـ ز ـ ژ |
| ـس ـ ـش ـ ـص ـ ـض | س ـ ش ـ ص ـ ض ـ ط ـ ظ |
| ـق ـ ـن ـ ـل | ق ـ ن ـ ل |

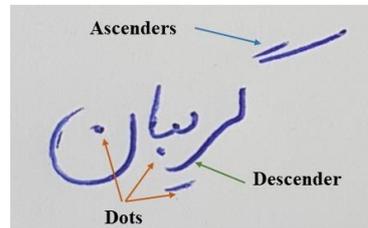

Fig. 3. The image of the word "گریبان" (Gariban which means **collar**).

- Many letters do not have a unique cursive shape. Even one individual may write the same letter in various shapes. For example, letters "س" (Sin), "ش" (Shin) (in each of their four forms), "ک" (Kaaf), "گ" (Gaaf), "ل" (Laam, particularly when combined with "ا" (Aa)), "م" (Mim), "ن" (Noon), "ی" (Y-e) and "ه" (H-e) are the most variable ones. As shown in Fig. 4, for example, the letter "ه" is written in two common shapes when it terminates a word, it also has two shapes when it comes before "ا" (Alef, in "ها", Haa) and is written in two shapes when it comes in the middle of a word (see Fig. 4.)

- There are also different shapes for some digits. Digits 4, 5, and 6 are the ones that may be written in completely distinct shapes as can be seen in Fig. 4.

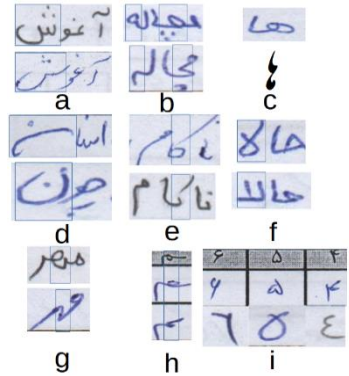

Fig. 4. Same letters and their cursive shape variations. a) The isolated form of the Letter "ش" (Shin), b) Two shapes of "ه" (final form of letter "ه", H-e) and "ج" (middle form of the letter "ج", Ch-e), c) Two shapes for the combination "ها" (Haa), d) Two shapes for isolated form of letter "ن" (Noon), e) Two shapes for "ک" (the initial form of letter "ک", Kaaf, f) Two shapes for the combination "لا" (Laa), g) Two shapes of "ه" (the middle form of the letter "ه", He), h) Two forms of "م" (the final form of the letter "م", Mim), i) The different cursive shapes of the digits "۶", "۵" and "۴" (Shesh, Panj, and Chehar, meaning 6, 5 and 4, respectively) from left to right.

In the following section, we will explain the Khayyam database in detail.

## 3. Dataset

### 3.1 Collecting Data

For gathering a handwriting dataset, we must fulfil the following requirements:

1. Generating various forms for gathering handwritten samples as diverse as possible to cover many potential shapes of the words, letters, and signs of the language. These forms must be designed to be easily parsed by a computer program.
2. Access to a community of responsible native writers who would like to fill out the forms.
3. Scanning the forms with acceptable quality
4. Having a computer program to extract the desired handwritten data, providing labelled data, and categorizing them to be useful for learning applications.

Our forms generally were filled by university and high school students voluntarily during in-class sections and at home. They had thirty minutes to fill them in class. The timing criteria were applied to provide us with handwriting samples in a more natural way.

Thanks to previous efforts like [14, 8], handwritten digits and letters data are sufficiently available for researchers. Besides, the letters and digits are limited to 10 and 34 classes, respectively, so the variability and also real-world applicability is lower than words. Hence, we concentrated more on gathering word images for the Khayyam dataset. Consequently, we focused on highly frequent words in Persian texts.

We designed three main types of forms. In Form 0 (Fig. 6), writers are requested to write five samples for digits and for each letter in their different forms on five pages. A few words also were included. In Form 1 (Fig. 7), we requested each writer to write about 450 frequently used words, extracted from official letters.

Form 1 also consists of words extracted from the literature books of all six grades of Iranian primary schools. The form includes 400 different words. This resource as our word pool fulfilled two purposes. First, we believe that the most reliable sources who may have concerns about the frequency of usage of every word in any language are those who are designing primary school books. This way our dataset remains as up-to-date as possible with the current situation of the language in the matter of word usage. Furthermore, as there is a huge interest in turning to paperless education in the educational community, systems trained by employing the most relevant data can be exploited together with some hardware such as digital papers to achieve this goal.

The third form (Form2) was a two-page form combined of letters, digits, numerical strings, and sentences. A few words also were included (Fig. 8 and Fig. 9).

All the forms were scanned using an EPSON L360 scanner with 300 dpi resolution. A hierarchy of all types of forms is shown in Fig. 5.

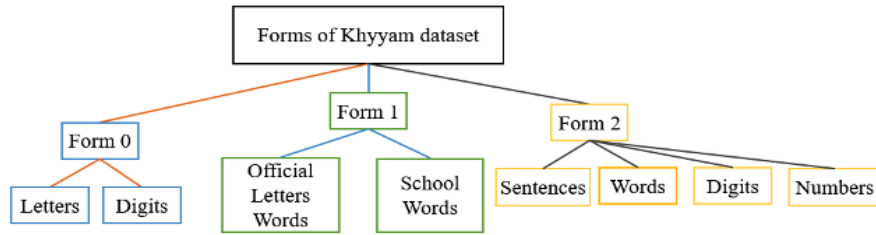

Fig. 5. A hierarchy of all the forms we generated to gather Persian handwritten samples.

Fig. 6. Page 1 of Form 0 for gathering the isolated form of the Persian letters. The other three pages were for gathering the Initial, Medial and final forms of the letters.

Fig. 7. Page 1 of Series 1 of Form 1 for gathering official words (page 2 is almost the same except for the writer information).

Fig. 8. Page 1 of Form 2 for gathering letters, digits and sentences (some words are included).

Fig. 9. Page 2 of Form 2 for gathering sentences.

### 3.1.1. Letters Dataset

Each of the 32 Persian letters can get one of two, three, or four possible distinct forms based on its position in a word. The possible forms are called isolated, initial, medial, and final.

As mentioned earlier, we gathered samples of the Persian letters in two types of paper forms (Fig. 6, Fig. 8.) To increase generality and to provide data for more than one size of text, the blank squares of the Form 0 are slightly bigger than the blank squares in the Form 2.

We expected that in bigger blank spaces people would write more freely resulting in bigger letters and in smaller places, they would take more caution and write smaller shapes of letters.

The Persian letters were gathered in four possible forms. On each page of Form 0, the writers were requested to provide 5 samples for each form of letter. In Fig. 10 we can see the difference between the samples in the two forms.

The dataset has 67000 handwritten samples for Persian letters; and 600 samples for each one of the four possible forms of each letter.

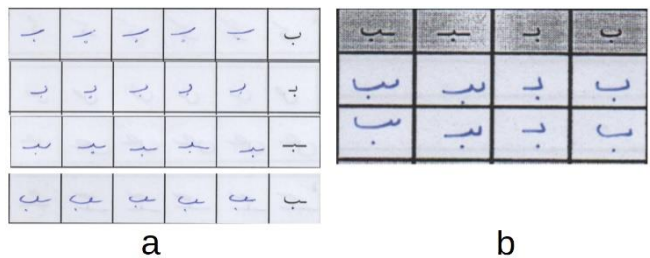

a  b

Fig. 10. Samples for all forms of letter "ب" (Be) a) In Form0. b) In Form2.

*Digit and Number Dataset*

Form 0 and Form 1 provided us with handwritten digits and numbers. The main challenge in collecting handwritten numbers are the occurrence of touching digits; consecutive digits connected to each other. In aggregate, the dataset has 6000 digit samples, with 600 samples per digit, and 1500 number samples (Fig. 11)

### 3.1.2. Sentence Dataset

Employing Form2 we collected sentence samples. The text dataset contains 250 distinct sentences. Sentences are randomly chosen, but a few considerations are considered. First, sentences are chosen in order to have a collection of words containing all the letters in alphabet with all their forms. Another consideration was to have the most reasonable distribution of letters to have a natural bias in frequency of usage towards common letters like "م" (Mim), "ا" (Alef), and "ن" (Noon).

Additionally, to make some less-common letters more present in the word dataset, we added extra word samples in Form 2 to compensate for the scarcity of some letters. For example, the words "آذر" (Azar: Fire), "مثلث" (Mosalas: triangle), "نظر" (Nazar: opinion), "ژاله" (Zhaleh: dew), "مچاله" (Mochaleh: Crumpled) were added to bring us extra samples of usages for letters "ذ" (Zal), "ث" (S-e), "ظ" (Z), "ژ" (Zh-e), "چ" (Ch-e), respectively. Fig. 12 shows some parts of the text.

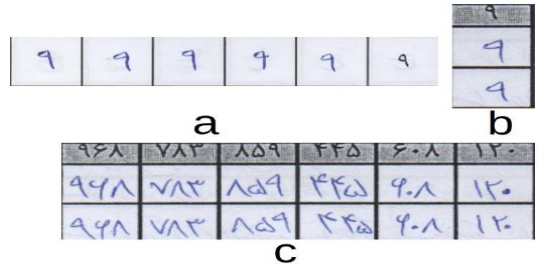

Fig. 11. a) Samples of digit 9 from Form0. b) The same samples gathered from Form2. c) The numerical strings from Form2.

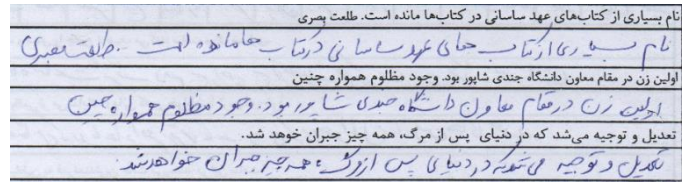

Fig. 12. A part of the text in the dataset.

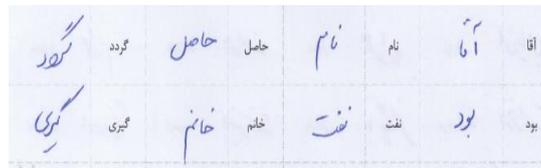

Fig. 13. Examples of words gathered using Form1. From top right to top left: "آقا" (Agha, meaning Mr.), "نام" (Naam, meaning name), "حاصل" (Haasel, meaning result). From bottom right to bottom left: "بود" (Bood, meaning was), "نفت" (Naft meaning oil), "خانم" (Khaanom, meaning Mrs.), "گیری" (Giri, the second part of many verbs in Persian meaning making, gaining, or doing, e.g. "تصمیم‌گیری" which means making decision)

### 3.1.3. Words Dataset

Primary school literature books are used as the main pool of words in the dataset. We extracted 400 distinct words from the books of all six grades in primary school. Using Form 1, 30,000 word samples (80 per word) were collected. Additionally, a form containing 450 distinct words, mostly official words, pronouns, and transitional words, was designed which also provided us with 18000 word image samples (40 per word, Fig. 13). Therefore, the Khayyam database contains 48000 word image samples. A related comparison with other public databases is presented in Table 2.

Table 2. Number of words in the datasets

| Dataset | Number of Words |
| --- | --- |
| Persian [8] | 71,000 |
| IAM [3] | 115,300 |
| IFN/ENIT(train + test) [7] | 42,700 |
| IFN/ENIT (Persian) [11] | 7,271 |
| PHCWT [17] | 3,600 |
| PBOK [6] | 27,073 |
| **Khayyam** | **48,000** |

### 3.1.4. Arithmetic Signs Dataset

The dataset also contains basic arithmetic signs. Primary school and university students were the main writers. The basic signs "+, −, ×, ÷, =, (, )" were collected. In Fig. 14 some samples of filled math signs are shown.

Table 3. Indices of letters of the Persian alphabet which the letter subdirectories are named according to.

| Letter | آ | ا | ب | پ | ت | ث | ج | چ | ح | خ | د | ذ | ر | ز | ژ |
|---|---|---|---|---|---|---|---|---|---|---|---|---|---|---|---|
| Index | 1 | 2 | 3 | 4 | 5 | 6 | 7 | 8 | 9 | 10 | 11 | 12 | 13 | 14 | 15 |
| Letter | س | ش | ص | ض | ط | ظ | ع | غ | ف | ق | ک | گ | ل | م | ن |
| Index | 16 | 17 | 18 | 19 | 20 | 21 | 22 | 23 | 24 | 25 | 26 | 27 | 28 | 29 | 30 |
| Letter | و | ه | ی | ئ | | | | | | | | | | | |
| Index | 31 | 32 | 33 | 34 | | | | | | | | | | | |

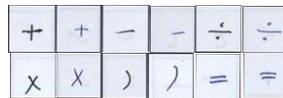

Fig. 14. Handwritten math signs

### 3.1.5. Filesystem Structure

The file structure of the dataset is tried to be as clear as possible. Fig. 15 illustrates the tree structure of the dataset files. There are subdirectories for digits, numbers, letters, words, sentences.

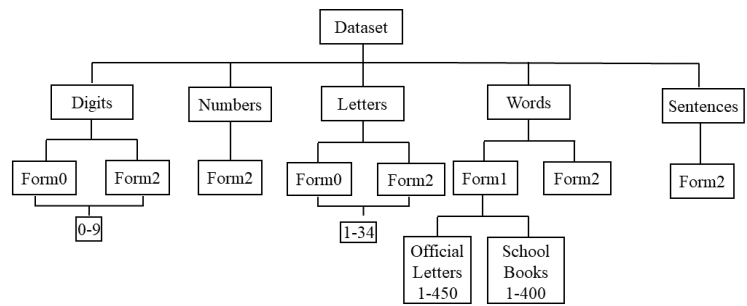

Fig. 15. File structure of the dataset

### 3.1.6. Datasets Statistics

To our knowledge, this chart is pretty close to the distribution of letter usage in the Persian language. A summary of the statistics of the whole dataset is also stated and compared to other available datasets in Table 4.

### 3.1.7. Preprocessing

Forms that distributed among writers, were scanned into a computer as TIFF and JPG images. A noise reduction, followed by a thresholding method produced the final binary images. In Fig. 16 the process of word extraction is illustrated.

Table 4. Statistics of the dataset.

| Data Type | Per sample | All |
|---|---|---|
| Words | 100 | 48000 |
| Letters | 600 | 67,000 |
| Digits | 600 | 6,000 |
| Numbers | 250 | 1,500 |
| Sentences | 130 | 3,200 |
| Each Arithmetic Sign Samples | 80 | 500 |

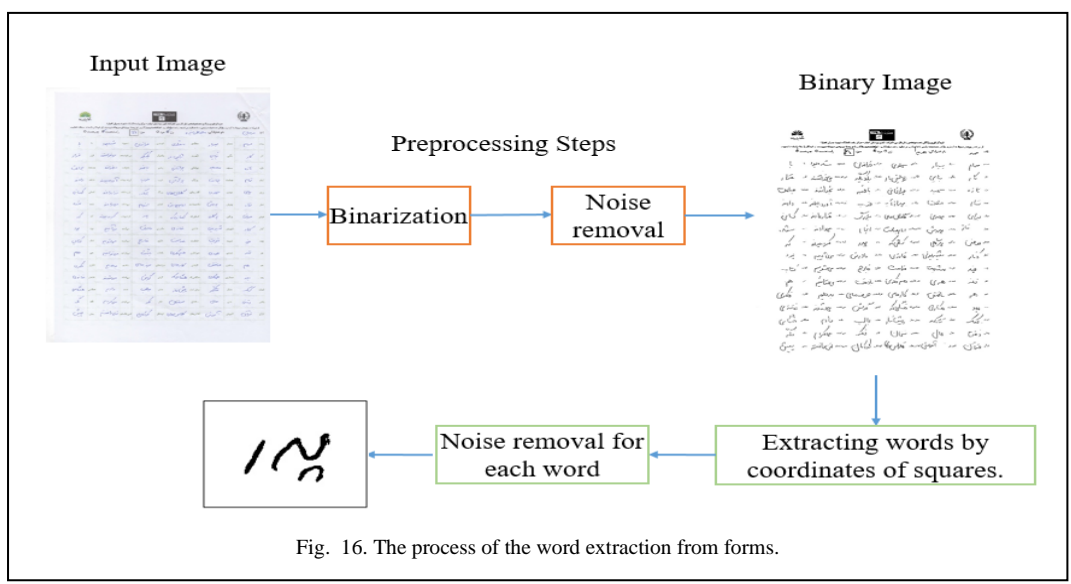

Fig. 16. The process of the word extraction from forms.

## 4. Experimental Results

To evaluate our dataset, we applied some of the important and popular machine learning methods to the dataset. For this purpose, training and test parts are produced. We allocated 90 and 10 percent to training and test, respectively and the KNN and SVM methods were applied to the dataset. In addition, a deep neural network was trained from the same configuration.

*4.1.  KNN*

We used the KNN method for digit and character classification. The histogram of the oriented gradient (HoG) of the images is extracted as the features used for KNN [21], then a KNN method is applied for classification. The pre-processing step, which is the most important part of the algorithm, consists of binarizing the input grey image using Otsu's thresholding, cropping and centring the binary image, and resizing the result to a 64x64 image. The 64x64 image will then be divided into 8x8 patches (Fig. 18). In each of the patches, the HoG of the pixels is calculated. Eight orientations were considered for the phase of the gradients (Fig. 17). Hence every patch is associated with an eight-dimensional vector. Therefore, the vector generated from contacting all these eight-dimensional vectors through the image will represent the whole image with its size. In the KNN algorithm after extracting this representative vector for each of the test images and calculating its Euclidean distance from the ones of the labelled training images, its class (label) will be identified as follows.

The results of the KNN method on the digit dataset of the Khayyam database are shown in Table 5. In brief, to indicate how a classifier confuses samples belonging to two classes with each other, the Confusion Matrix in Table 7 is produced. For example, clearly in the confusion matrix, the classifier has confused the digits 2 ('۲') and 3 ('۳') many times since they are similar in shape.

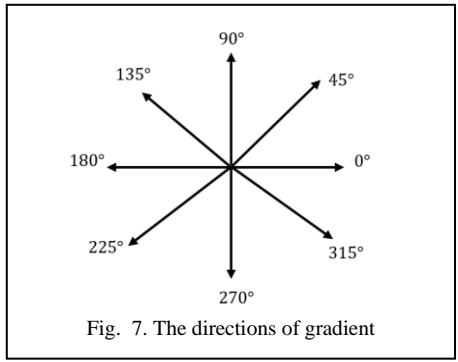

Fig. 7. The directions of gradient

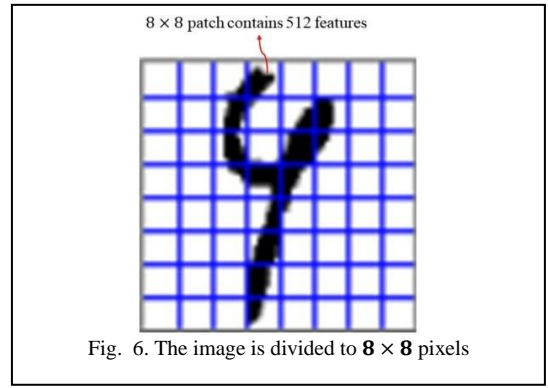

Fig. 6. The image is divided to $8 \times 8$ pixels

Table 5. Experimental results for digits

| Method | Accuracy |
|---|---|
| KNN (k=3) | 92.33 |
| KNN (k=5) | 92.10 |
| KNN (k=7) | 92.56 |

*4.2. Support Vector Machine*

The support vector machine (SVM) method is one of the most important parametric classification methods. Although the SVM method is mainly applied to two-class problems, it can be used in multi-class problems with a few modifications. Due to the training imbalance in the one-versus-the-rest approach [22] (e.g. we needed nine negative classes versus just one positive class in digit recognition), we used the one-versus-one approach in which classes are trained and every pair of classes is considered as positive and negative, separately. We evaluated the performance of this method in the classification of our digits and letters datasets. To achieve this goal, the HoG feature vectors in the previous section were used as input vectors for SVM. The results are shown in Table 6.

Table 6. Experimental results of the SVM method on digits and letter.

| Class | Accuracy |
|---|---|
| Letters | 90.54 |
| Digits | 92.65 |

Table 7. Confusion matrix in Khayyam dataset (KNN, K=7).

| | | Recognized as | | | | | | | | | |
|---|---|---|---|---|---|---|---|---|---|---|---|
| | Digits | 0 | 1 | 2 | 3 | 4 | 5 | 6 | 7 | 8 | 9 |
| | 0 | 57 | 0 | 0 | 1 | 0 | 1 | 0 | 0 | 0 | 0 |
| | 1 | 0 | 88 | 0 | 0 | 0 | 0 | 0 | 0 | 0 | 3 |
| | 2 | 0 | 1 | 77 | 10 | 0 | 0 | 2 | 2 | 0 | 0 |
| Input digit | 3 | 0 | 0 | 9 | 79 | 2 | 0 | 0 | 1 | 0 | 0 |
| | 4 | 0 | 3 | 3 | 3 | 80 | 0 | 1 | 0 | 0 | 0 |
| | 5 | 1 | 0 | 0 | 0 | 0 | 90 | 0 | 0 | 1 | 0 |
| | 6 | 0 | 2 | 1 | 0 | 1 | 0 | 78 | 1 | 0 | 3 |
| | 7 | 0 | 2 | 3 | 1 | 0 | 0 | 0 | 86 | 0 | 0 |
| | 8 | 0 | 1 | 0 | 0 | 0 | 1 | 0 | 0 | 87 | 0 |
| | 9 | 0 | 1 | 0 | 0 | 0 | 0 | 4 | 0 | 0 | 87 |

## 4.3. Deep Neural Networks

Deep Learning has had a huge impact on handwriting recognition applications in different languages as well as on Persian handwritten recognition [23, 24, 25, 26, 27, 28]. Here deep models are trained using the words, letters, and digits in the Khayyam dataset. The model consists of convolutional (CNN), bidirectional LSTM (BLSTM) layers with CTC loss function [29].

Convolutional neural networks generally use image data for classification purposes [30, 31]. Also, bidirectional LSTM layers are used for evaluating sequence data [32] and they are used in handwriting recognition tasks [33, 34]. The CNN-LSTM network model can be applied to handwriting recognition tasks. CNN layers can learn local correlations along with hierarchical correlations, based on the learned local features, LSTM layers can learn long-term dependencies. This combination overcomes the weaknesses of the two networks [35]. Besides, the connectionist temporal classification (CTC) loss function was introduced for sequence labelling without a segmentation step [36].

Here a deep model with CNN and RNN layers for word recognition is used. Using the CTC layer the model doesn't need segmentation steps. The character accuracy, Top1, and Top5 parameters in Table 10 and Table 11 are calculated from the below formulas:

$$accuracy = 100 \times (1 - \frac{insertions + deletions + substitutions}{total\ length\ of\ all\ the\ words\ in\ test\ set}) \quad (1)$$

$$Top1 = 100 \times (\frac{\#\ of\ samples\ that\ their\ lebels\ are\ the\ highest\ score's\ model\ choice}{\#\ of\ All\ input\ test\ data}) \quad (2)$$

$$Top5 = 100 \times (\frac{\#\ of\ samples\ that\ their\ lebels\ are\ the\ first\ 5\ highest\ score's\ model\ choice}{\#\ of\ All\ input\ test\ data}) \quad (3)$$

The results of the applied model on the digits, letters, and words are shown in Table 8, Table 9, Table 10, and Table 11, respectively.

Table 8. Experimental results for digits

| Method | Accuracy |
|---|---|
| CNN(6 layers) + BLSTM(2layers) | 97.37 |

Table 6  Experimental results for letters

| Method | Accuracy |
|---|---|
| CNN(7 layers) + BLSTM(2layers) | 94.96 |

Table 10 Experimental results for official letter's words

| Method | Character Accuracy | Top1 | Top5 |
|---|---|---|---|
| CNN(9 layers) + BLSTM(2layers) | 94.94 | 94.03 | 97.25 |

Table 11 Experimental results for school's words

| Method | Character Accuracy | Top1 | Top5 |
|---|---|---|---|
| CNN(9 layers) + BLSTM(2layers) | 93.42 | 95.65 | 98.95 |

## 5. Conclusion

In this paper, the Khayyam dataset for unconstrained Persian handwriting is introduced. The statistics of the dataset were described in detail. The data are collected both male and female native writers. It has ground truths for all the data types such as sentences, words, digits, and letters. Thanks to different sizes, we can claim that the variability of data is high. Due to the availability of sufficient labelled data for digits and letters from other research, we mainly concentrated on gathering word samples. To have a comprehensive collection of the most basic and important words, a large portion of them is extracted from literature books taught in primary schools. We tried to keep the distribution of the usage of Persian letters in the word and sentence datasets as close as possible to the real distribution of them in the language. The database contains 48000 labelled word images which are a good resource for training trending machine learning and deep learning models. To confirm this claim, we performed KNN and SVM methods and deep neural network models on the dataset and the results are reported. The dataset can be used in many fields of document processing such as word recognition, word segmentation, word spotting, word separation, and number recognition.

## 6.    Acknowledgment

The authors of this paper would like to thank all the students at the Shahid Chamran University of Ahvaz, Iran who generously helped us by filling out the paper forms. Definitely, without their assistance, this project would not have been accomplished.